\documentclass{article}

\usepackage{times}
\usepackage{epsfig}
\usepackage{graphicx}
\usepackage{amsmath}
\usepackage{amssymb}
\usepackage{tikz}
\usepackage{subcaption}

\usepackage{hyperref}

\usepackage[accepted]{icml2021}

\icmltitlerunning{Statistically Significant Stopping of Neural Network Training}

\begin{document}




\twocolumn[
\icmltitle{Statistically Significant Stopping of Neural Network Training}



\icmlsetsymbol{equal}{*}

\begin{icmlauthorlist}
\icmlauthor{J. K. Terry}{umd,swarm}
\icmlauthor{Mario Jayakumar}{umd,equal}
\icmlauthor{Kusal De Alwis}{umd,equal}
\end{icmlauthorlist}

\icmlaffiliation{swarm}{Swarm Labs}
\icmlaffiliation{umd}{Department of Computer Science, University of Maryland, College Park}

\icmlcorrespondingauthor{J. K. Terry}{jkterry@umd.edu}

\icmlkeywords{Deep Learning, Classifier, Optimization}

\vskip 0.3in
]



\printAffiliationsAndNotice{\icmlEqualContribution} 


\section{Disclaimer}

The specific statistical test and methodology we use in this paper is not valid as it relies on accepting a null hypothesis, which cannot be validly done. While we believe that our fundamental intuition of neural net validation accuracy differences during training being normally distributed indicating an optimal stopping point as valid, we are unable to create a successful implementation of this test. To implement such a test, a method is required to determine whether a given distribution is normally distributed, or at least close to normal. Statistical significance tests are unable to do this as all statistical significance tests for normality rely on a null hypothesis that the provided distribution is normal. As is the case with null hypothesis statistical tests, it is possible to claim with a certain level of confidence that a population is not normal solely based on the normality of random samples. However, claiming that the population is truly normal is much more difficult as the entire population would need to be measured: even a slight deviation from a normal distribution can make a population no longer "normal" \cite{kluger2001error,schneider2015null}. Therefore, methods other than statistical significance tests must be used. As shown by  \cite{jarque1987test, decarlo1997meaning, jones1969skewness}, kurtosis and skew tests are likely to be useful measurements for normality, however we were unable to successfully implement these. Moreover, our tuning of the slackProp hyperparameter in our implementation was not methodologically valid because it allowed precise stopping points to be chosen for specific datasets given knowledge that may not be present in application.

Useful contributions of this paper and repository for future work is a large corpus of neural network training data as well as a novel approach to the early stopping of neural network training. We have trained 10 different models on the CIFAR10 dataset and AlexNet on the ImageNet dataset, recording training and validation data for each epoch, with 10 training sessions each. This equates to 100 recorded training sessions for CIFAR10 across 10 models, and 10 recorded training sessions for ImageNet. We have additionally provided training data for learning rate schedule experiments for ResNet101 and GoogLeNet. This data was expensive to obtain, and can help future researchers avoid further compute expenses. This dataset can be used for the analysis of neural network training patterns, and further research into stopping methods based on neural network accuracy during training. Additionally, our implemented ASWS stopping method indicates good results compared to existing stopping methods (with the tuning of the slackProp hyperparameter), but does not accurately follow our original hypothesis of testing for normally distributed testing accuracy. 

\begin{abstract}

The general approach taken when training deep learning classifiers is to save the parameters after every few iterations, train until either a human observer or a simple metric-based heuristic decides the network isn't learning anymore, and then backtrack and pick the saved parameters with the best validation accuracy. Simple methods are used to determine if a neural network isn't learning anymore because, as long as it's well after the optimal values are found, the condition doesn't impact the final accuracy of the model. However from a runtime perspective, this is of great significance to the many cases where numerous neural networks are trained simultaneously (e.g. hyper-parameter tuning). Motivated by this, we introduce a statistical significance test to determine if a neural network has stopped learning. This stopping criterion appears to represent a happy medium compared to other popular stopping criterions, achieving comparable accuracy to the criterions that achieve the highest final accuracies in 77\% or fewer epochs, while the criterions which stop sooner do so with an appreciable loss to final accuracy. Additionally, we use this as the basis of a new learning rate scheduler, removing the need to manually choose learning rate schedules and acting as a quasi-line search,  achieving superior or comparable empirical performance to existing methods.
\end{abstract}

\section{Introduction}

\begin{figure}[t]
    \centering
    \scalebox{0.5}{\input{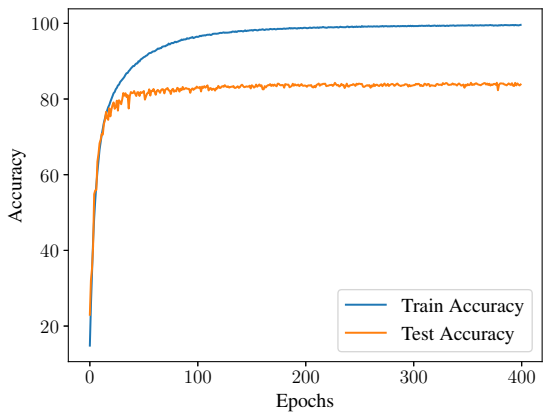}}
    \caption{Train and Test accuracy for training Alexnet on the ImageNet dataset, optimized with SGD (LR=.1).}
    \label{TrainVsTestCurve}
\end{figure}

Neural networks are a powerful tool for solving classification problems. This has led to their ubiquitous use throughout all areas of academia and industry. A major component of their power stems from their ability to represent any function, but this comes with a downside. When training a neural network, only a subset of the true distribution of the data for the task is ever available for training, and minimizing loss w.r.t this sample is the objective of training. Typically a point will be reached after which the neural networks parameters begin to fit noise in the training data so effectively that they perform worse on unseen data from the broader distribution (neural networks overfit, resulting in worse generalization), or the neural networks stop being able to meaningfully improve its loss or accuracy even on the training set. Therefore, for generalization or compute time reasons neural networks need to be stopped at some point before they truly finish optimizing the loss function. The most common way to address this is to split the available data into two sets. One set is used for training, and a second smaller set (which this paper will refer to as the testing set) is used to evaluate the loss and/or accuracy (this paper collectively refers to the two as "error") of the neural network when inferencing on data that it hasn't been trained on, throughout the training process. On plots of the error vs training iteration, the testing curve will eventually split from the training curve; an example is shown in \autoref{TrainVsTestCurve}. The optimal place to stop the neural network's training is thus when it has the minimum error on the testing set. This also serves as a form of implicit regularization as shown by \cite{zhang2016understanding}.

If you train for a very long time, the point of best performance will be obvious and you can pick those saved parameters from training. However, determining a termination condition to know when to stop is not a straightforward process. An obvious approach is to do this manually, and when a person is doing a single simple run this is fine. However in the contexts of AutoML and hyperparameter searches, a programmatically defined condition must be used. These are always heuristics, a wide variety of which are used in practice. A popular tutorial on early stopping methods, \cite{brownlee_2020}, recommends using the following stopping conditions:

\begin{itemize}
  \item No change in test accuracy over a given number of epochs, which we will refer to as the patience stopping method.
  \item An absolute change in accuracy or loss.
  \item A decrease in highest test accuracy observed over a given number of epochs, which we refer to as the performance stopping method.
  \item An average change in test accuracy over a given number of epochs, which we will refer to as the average change based early stopping method.
\end{itemize}

Similarly TensorFlow \cite{abadi2016tensorflow}, through tf.keras, has built in early stopping functionality, that stops after an error metric changes by less than a defined value a certain number of times \cite{community_2020}.


In the context of AutoML, where early stopping matters the most, you're generally training batches of neural networks at once. In these cases, under-performing networks in a batch are ``pruned'' (stopped) via techniques which notably include \emph{median stopping rule} \cite{46180} and \emph{HyperBand} \cite{li2017hyperband}. These techniques, however, do not tell if a run is done or not like the techniques this paper is concerned with do, they just say if the run is likely to be for a useful hyperparameter. Previously described standard termination conditions are still used to finally stop the ``surviving`` runs in these cases.

A related line of inquiry to our work is the field of ``optimal stopping.'' This is a mature field of math and statistics that studies when to optimally stop an iterative process, and the optimal stopping of neural network training is ultimately a special case in this line of research. The existing optimal stopping literature has largely focused on problems in the spirit of when to finally accept offers on a house from a procession of potential buyers, or when to stop trying out new secretaries or assistants \citep{ferguson_2020}. The field as produced proofs of numerous optimal strategies for general problems, however these works generally assume that the procession of new data points is I.I.D. random or similar. This is a very invalid assumption for the case of neural network loss of accuracy curves. While outside of the theoretical optimal stopping literature which has guarantees, ML methods have recently been used for the optimal stopping problem in random cases \citep{becker2019deep}, or for pruning the training of a neural network as part of a set during hyperparameter tuning \cite{dai2019bayesian} (though \emph{not} for individual training runs).

Early stopping has not been exhaustively explored because it doesn't impact the performance of the trained model. The selection of termination conditions can significantly impact training time, and therefore reduce the cost and environment impact of training in deep learning. Given the vast amounts of resources devoted to training neural networks, these are meaningful considerations. Moreover, these heuristics are just that--heuristics. They don't attempt to offer a substantive answer to the question ``has my neural network stopped learning?,'' or equivalently ``has my neural network started fitting to noise in the training data instead of general trends?'' Motivated by all this, we introduce a statistical significance test to determine if a neural network has stopped learning, by only looking at the testing set accuracy curve. Our test is an extension of the Shapiro-Wilk test, and as such we've creatively named it Augmented Shapiro-Wilk Stopping (``ASWS``). 
ASWS appears to represent something of a happy medium in the space of stopping conditions. It stops in 77\% or less steps than all popular conditions but two without a meaningful loss in final accuracy. The two conditions that stop sooner than the ASWS do so at the expense of 2-4\% final accuracy even with tuned hyperparameters (e.g. they stop too soon). We additionally are able to use this to create an automatic LR-scheduler akin to a line search, that converges to comparable accuracies to conventional methods in the same, or often fewer, number of epochs.

\section{Background}
Here we lay out a few concepts key to this paper that may be unfamiliar to some members of the deep learning community.

\textbf{Shapiro-Wilk Test}:

Introduced by \cite{shapiro1965analysis}, the Shapiro-Wilk Test determines the probability that a sample of data points was drawn from a normal distribution. It is the most powerful normality test, demonstrated by \cite{razali2011power}, and notably for this paper, it is agnostic to the mean.

\textbf{Single Sample T-Test}:

The single sample t-test, defined formally in \cite{student1908probable}, determines the probability that a sample of data points was drawn from a distribution with a mean other than a specified one.

\textbf{Clipped Exponential Smoothing}:

Exponential smoothing is a method for smoothing time series data, via the formula $\mathsf{smooth}_i = \frac{\gamma \cdot \mathsf{smooth}_{i-1} + T_i}{w_i}$ where $T_i$ is the original time series, $0 < \gamma < 1$ is the smoothing factor, $w_i$ is defined recursively as $w_i \, = \gamma \cdot w_{i-1} + 1$ and $\mathsf{smooth}_0$ is initialized to $T_0$ and $w_0$ is initialized to 1. Clipping changes this to only exponentially smooth up till a defined clipping point $c$, which removes the possibility of tail effects having adverse consequences and slightly decreases computation time, and only has the restriction that $c < |T|$. After this point, the smoothing formula is now defined as  $\mathsf{smooth}_i = \frac{\beta_{j}}{w_c}$. The term $\beta_{j}$ is also recursively defined as $\beta_{j}=\beta_{j-1} \cdot \gamma + T_j$ until the base case $\beta_{i-c} = 0$. The exponential smoothing family of methods is reviewed in \cite{gardner1985exponential}, and is the most simple and popular method of smoothing time series in the physical sciences 

\section{Augmented Shapiro-Wilk Stopping}

\subsection{Intuition}
\label{sec:intuition}

While training, accuracy on the test dataset will generally be increasing, with a high degree of noise from random sampling of the data, and numeric errors amongst other sources. When the variations in the test accuracy curve become purely noise and their mean is zero, then you can be fairly confident that learning has stopped (as far as improvements to the validation set accuracy curve are concerned). Per the central limit theorem, when these variations are random they will also be normally distributed. This creates two questions well suited for the statistical significance tests described in the previous section. The Shapiro-Wilk test can tell you if the recent accuracy values are normally distributed, and the simple sample t-test will tell you if they have zero mean.

The problem with this is the nature of the noise during training. If the noise of an the error curve you're looking at is too extreme, then any changes will become washed out. Furthermore, the noise seen is very dependent on the neural networks loss landscape, meaning that the variations are not I.I.D (Independent and Identically Distributed random variables) on small time scales. These factors make any statistical analysis very challenging. However, we've found that by combining three mitigations to the problem of noise together, good results can still be achieved:

\begin{itemize}

  \item Smooth error curves. Sensible amounts of smoothing can't meaningfully change the macro-trends we're looking at here, and removing the overall level of noise will make it much easier to find.
  \item Only look at the testing set accuracy curve. Overcoming the much greater noise and non-I.I.D. characteristics of loss curves remains a future work.
  \item Check for both 0 mean and normal distribution at once. Both should be true when a neural network is done training, and testing for both dramatically decreases false positive rates.

\end{itemize}

The key point of this test is the remarkable statistical power of the Shapiro-Wilk test, and from our perspective everything else is a modification of this core test to address the idiosyncrasies of neural network error curves. As such, we term our algorithm for determining if a neural network has stopped \emph{Augmented Shapiro-Wilk Stopping} (or ``ASWS''). 

\subsection{The Full Algorithm}

Consider a neural network that has trained for $k$ epochs, and we have a list $testAcc$ of length $k$, where $testAcc[i]$ has the test accuracy of the neural network at epoch $i$. We additionally define $\Delta testAcc$ as the finite difference of $testAcc$, as computed by Numpy's gradient function. We additionally define $\alpha$ as the desired probability that learning has stopped (we used .97), $\gamma$ and $c$ relate to parameters for exponential smoothing, $n$ is the sample size we use for our single sampled t-test and Shapiro-Wilk test. It is necessary that $n \leq k$ in order to perform the ASWS test, since we would like to isolate training noise that occurs initially from noise that occurs later. Once $k > n$, we perform a sliding window over $testAcc$ where each window is of size $n$. These windows form the input to the statistical tests, and the number of windows generated will be $k-n$. Furthermore, this will generate $k-n$ pairs of Shapiro-Wilk test and single sample t-test results, over $testAcc$. Of course, not all these windows will be equally relevant: later windows will be more representative of the model's current performance while training. Consequentially, the statistical tests performed on later windows will be more salient. We define $slack$, where $k-n < slack$, as how many of the later statistical test results we use to determine when to stop. We additionally define $slackProp$, for $0 < slackProp \leq 1$, as what fraction of those later results must be statistically significant. More formally, we check the last $slack$ pairs of Shapiro-Wilk and single sample t-test results, and stop learning if at least $slack \cdot slackProp$ of them simultaneously surpass $\alpha$. The ASWS stopping algorithm is then as described in \autoref{ASWSAlgo}.

\begin{algorithm}[tb]
\caption{The ASWS Stopping Algorithm}
\label{ASWSAlgo}
\begin{algorithmic}
\STATE $\gamma$, $c$ = $\mathsf{Exponential Smoothing Parameters}$\\
\STATE $B[i]$ = $\mathsf{testAcc}[i:i+n]$ for $i \in [0, k-n]$\\
\STATE $\Delta B[i]$ = $\Delta \mathsf{testAcc}[i:i+n]$ for $i \in [0, k-n]$\\
\STATE $\mathsf{Smooth} \leftarrow$ $\mathsf{ClippedExponentialSmoothing}$($\mathsf{testAcc}$, $\gamma$, $c$) 
\FOR{$i \leftarrow 0$ {\bfseries to} $(k-n)$}

    \STATE $TTest[i] \leftarrow$ $\mathsf{SingleSampleTTest}$($\Delta B[i]$, $\mathsf{center}$=0)\\
    \STATE $Shapiro[i] \leftarrow$ $\mathsf{Shapiro}$($Smooth[i]$)\\
\ENDFOR

\STATE $stopReq \leftarrow 0$

\FOR {$i \leftarrow (k-n-slack)$ {\bfseries to} $(k-n)$} 
    \IF{$TTest[i] > \alpha$  and $Shapiro[i] > \alpha$}
        \STATE $stopReq \leftarrow stopReq + 1$\\
    \ENDIF
\ENDFOR
\IF{$stopReq>slackProp \cdot slack$}
    \STATE return True \%stop training\\
\ENDIF
\end{algorithmic}
\end{algorithm}

\subsection{Hyper-parameters}

This method has 5 hyperparameters: $n$, $\gamma$, $c$, $slack$ and $slackProp$. 


We find that $c$ does not have a large significant effect on training, so we use a value of 15 throughout. The values of $n$, $\gamma$ and $slackProp$ have varying effects on when stopping occurs, so we consider those the tunable hyperparameters for ASWS.

The confidence of the Shapiro and T-Test, $\alpha$, is another possible hyper-parameter but not one that we changed per model. Regardless of what neural network is being trained, the normality of a sample and its centeredness about 0 should remain consistent. Therefore, we chose to use a constant $\alpha=0.97$ for evaluation. 

\section{Experimental Evaluation}

The results of all experiments, the code for them, and the code to easily use the ASWS is available at \url{https://github.com/justinkterry/ASWS}.

\subsection{Methodology}

We trained 10 trials each of AlexNet \cite{NIPS2012_4824}, GoogLeNet \cite{szegedy2015going}, ResNet34, ResNet50, ResNet101 \cite{he2016deep}, VGG11, VGG16, VGG19 \cite{simonyan2014very} and 2 fully connected networks, fc1 (2 hidden layers containing 400 then 300 units), and fc2 (3 hidden layers containing 700, then 600, then 500 units) on the CIFAR10 data-set \cite{krizhevsky2009learning} with PyTorch \cite{paszke2019pytorch} to establish curves to evaluate our early stopping method with. Our training implementation was based on the repository provided by \cite{kuangliu}. We used plain SGD, with a learning rate of 0.1 and momentum of 0, as the optimizer for these baselines, because it's the most popularly used optimizer and the most challenging for our method---advanced momentum methods dramatically decrease periodic self correlation by not oscillating in the same region \cite{sutskever2013importance}. We train for 400 epochs every time to ensure we have training curves for each network that goes well past the point of being fully trained. Even thought it's standard, we don't use learning rate scheduling for these tests. This is because a neural network which has has ceased to improve will generally start learning again when the learning rate is decreased even it's previously stopped, which isn't ideal when trying to evaluate a stopping method alone. In practice, learning rate scheduling can be reconciled with ASWS by using it to indicate when the learning rate should be decreased, something we explore in \autoref{sec:lr_scheduling}.


There are four baseline early stopping methods we compare against, which are briefly outlined in the intro. The performance stopping method stops training if the test accuracy has not reached a new maximum in $\mathsf{K}=60$ epochs. The patience stopping method is to stop training if the test accuracy has decreased in $\mathsf{patience}$ consecutive epochs. The minimum difference stopping method is to stop training if the change in test accuracy has not improved by some minimum delta for $\mathsf{patience}$ consecutive epochs. Finally, the average difference stopping method stops training if the average of the differences in test accuracy over the last $\mathsf{window}$ epochs falls below some minimum delta. The optimal hyperparameters for these stopping methods were determined by a grid search which maximized average test accuracy across models, with search values as outlined in \autoref{HeuristicGridSearchValues}. Optimal hyperparameters are included in \autoref{StandardStopParams}. As is done in practice, we evaluate the model's final test accuracy as the highest test accuracy that the model achieved while training before the stopping point. Each model was trained 10 times, and results were averaged over them.



\subsection{Results}

\begin{figure*}
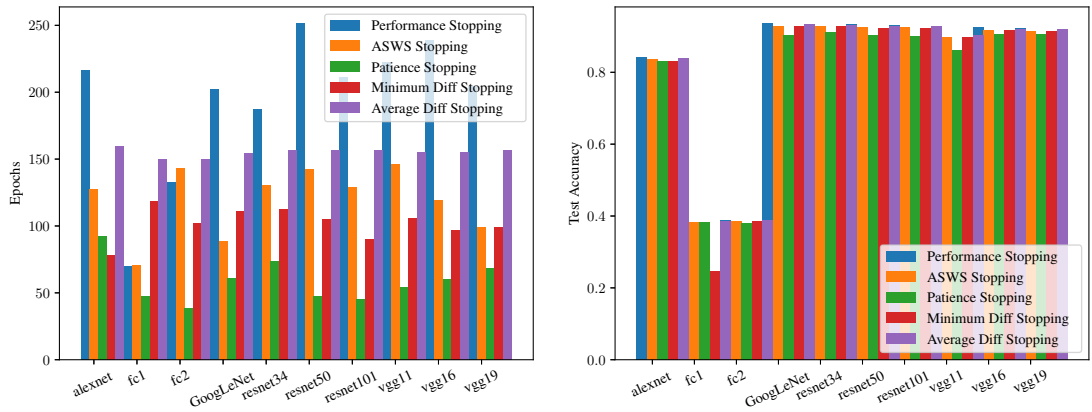

\vspace*{-1.9 in}
\begin{subfigure}{0.5\textwidth}

    \centering
    \scalebox{0.5}{\input{figures/ASWTStandardComp4.pgf}}

\end{subfigure}%
\hspace*{-0.5in}
\begin{subfigure}{0.5\textwidth}
    \centering
    \scalebox{0.5}{\input{figures/ASWTStandardCompByAcc4.pgf}}
    
    \end{subfigure}
        \caption{This plot compares the stopping epochs and final accuracy for various models trained on the CIFAR-10 dataset with the ASWS stopping method and other established stopping methods. Each model was trained 10 times for at most 400 epochs and averaged. The selected ASWS stopping points were determined by hyper-parameters which were optimized for that model. This shows that the ASWS is able to stop dramatically sooner than the performance method across a diverse set of neural networks. While stopping methods such as Patience stopping and Minimum Difference stopping tend to stop earlier than ASWS stopping, their effect on test accuracy is not as high. This shows that ASWS achieves a performance that is midway between the high performance of the standard methods and the aggressive stopping of the patience and minimum difference methods.}
    \label{ASWSStandardComp}
\end{figure*}

Our results, shown in \autoref{ASWSStandardComp} demonstrate that the Augmented Shapiro Wilk Stopping can achieve comparable test accuracy to standard stopping methods while saving a significant number of training epochs.

\begin{figure*}
    \vspace*{-0.9in}
    \hspace*{-0.6in}
    \centering
    \scalebox{0.7}{\input{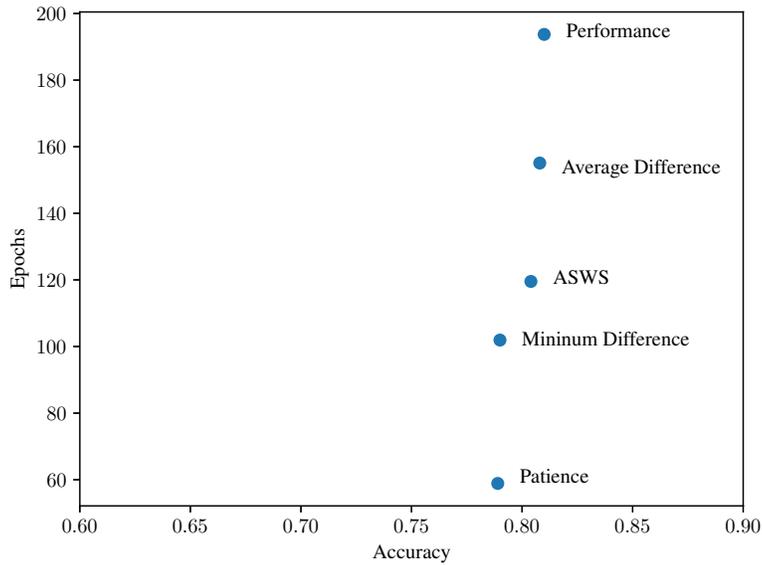}}
    \caption{This plot compares the average stopping epoch and accuracy for various stopping methods. Minimum difference and patience stopping methods stop the earliest, but have lower accuracies compared to ASWS, Average Diff and Performance stopping. ASWS has similar accuracies to Average difference and performance, but takes close to 60\% of the Performance method training time.}
    \label{fig:aswsepochvsaccplot}
\end{figure*}

The hyperparameters we used for ASWS were obtained via grid search, with search space shown in \autoref{GridSearchValues}, over the smoothing factor $\gamma$, the sample size $n$, and the slack proportion $slackProp$ over each model. The goal of this search was to minimize the average stopping point, without having an adverse impact on final model accuracy (we restricted it to 0.5\% less than the performance stopping method). The optimal hyperparameters we found are shown in \autoref{OptimizedHypers}.

The only stopping methods able to consistently achieve higher test accuracy, when compared to our ASWS method, are the performance stopping method and the average difference stopping method. The similar accuracy performance is shown in \autoref{fig:aswsepochvsaccplot} with ASWS, Average difference and Performance stopping having very similar accuracies; simultaneously, the ASWS stopping method stop within about 60\% as many epochs as Performance stopping. While the difference in test accuracy here is within 0.5\%, \autoref{ASWSStandardComp} shows the large disparity in stopping epoch, with our ASWS method achieving significantly earlier stopping. The minimum difference stopping method performs similarly to our ASWS method, though it tends to to stop slightly earlier than ASWS at the expense of test accuracy across models. Similarly, the patience stopping method stops much earlier than our method, but also has a lower accuracy across models. In terms of aggressiveness of stopping, ASWS falls in the middle of these common methods, achieving the accuracy of the slower methods in far fewer steps.
    
\begin{table}
\begin{center}
\begin{tabular}{ |c|c|c|c| } 
 \hline
    \textbf{Model} & \textbf{$\gamma$} & \textbf{n} & \textbf{slackProp} \\
 \hline
    AlexNet & 0 & 13 & 0.95 \\
 \hline
    fc1 & 0.1 & 5 & 0.132 \\
 \hline
    fc2 & 0.775 & 13 & 0.132 \\
 \hline
    GoogLeNet & 0.213 & 19 & 0.95 \\
 \hline
    ResNet34 & 0.55 & 17 & 0.868 \\
 \hline
    ResNet50 & 0.213 & 19 & 0.459 \\
 \hline 
    ResNet101 & 0.1 & 15 & 0.95 \\ 
 \hline
    VGG11 & 0.663 & 17 & 0.95 \\
 \hline
    VGG16 & 0.888 & 17 & 0.705\\
 \hline
    VGG19 & 0.1 & 17 & 0.623 \\
 \hline
\end{tabular}
\end{center}
\caption{Optimized hyper-parameters found for each model via grid search.}
\label{OptimizedHypers}
\end{table}

\begin{figure*}
    \centering
    \scalebox{0.7}{\input{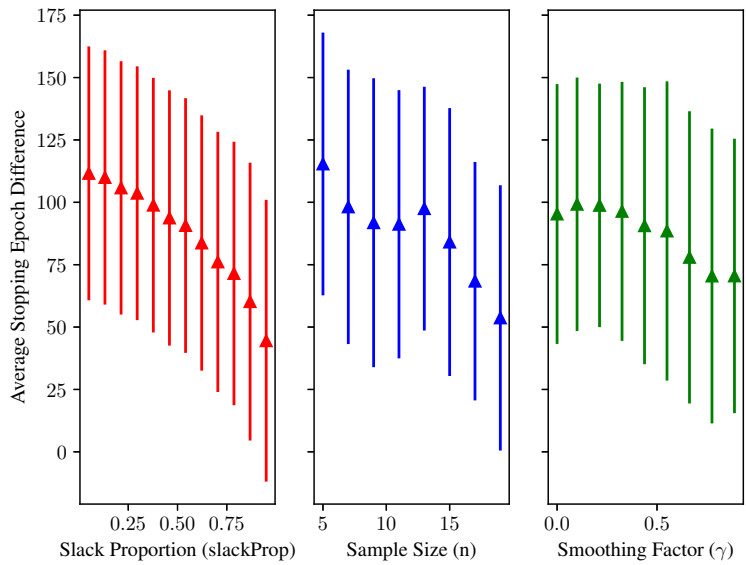}}
    \caption{Effect of hyper-parameters on the stopping epoch recommended by the ASWS. A grid search was performed using 10 training runs all models. The mean difference between the performance stopping point and the ASWS algorithm stopping point is shown for each tested hyper-parameter values (listed in \autoref{GridSearchValues}). Changes in one hyperparameter were averaged over all other searched hyperparameters, across all models as stated in \autoref{OptimizedHypers}. The marker represents the average stopping epoch difference when the respective hyperparameter is fixed. The error bars represent the standard deviation of stopping epoch difference.}
    \label{fig:hyperparameterseries}
\end{figure*}

Understanding the impact of these hyperparameters is important to practical usage of our method. To better understand the impact, we generated \autoref{fig:hyperparameterseries}. This plot shows the average performance impact of changing one hyperparameter, over all other tested hyperparameters. Both the plot and hyperparameter table show a positive result---the smoothing factor $\gamma$ doesn't have a significant impact on model performance, and tends to have a fairly consistent effect when compared to the other hyperparameters. Additionally, the effect of using no smoothing, when $\gamma=0$, shows that smoothing tends to cause earlier stopping, but then stops later. This implies that using no smoothing is possible but only AlexNet is optimal when no smoothing is used. Overall, the effect of $\gamma$ is consistent within the ranges of 0.1 and 0.5. This means no tuning is required in practice. The optimal sample size $n$ is intuitively primarily dictated by roughly how long the model should be trained for, and relatedly how many parameters it has, something born out in the data. If you just do a simple linear regression between the two, the r value is 0.81. This means that the impact of the hyperparameter is predictable, and similar models shouldn't require different values. Consequentially, tuning it should not be a significant practical problem. We unfortunately do not have such a way to avoid tuning the slack proportion though, and leave this as a future work. 
 
\begin{figure*}
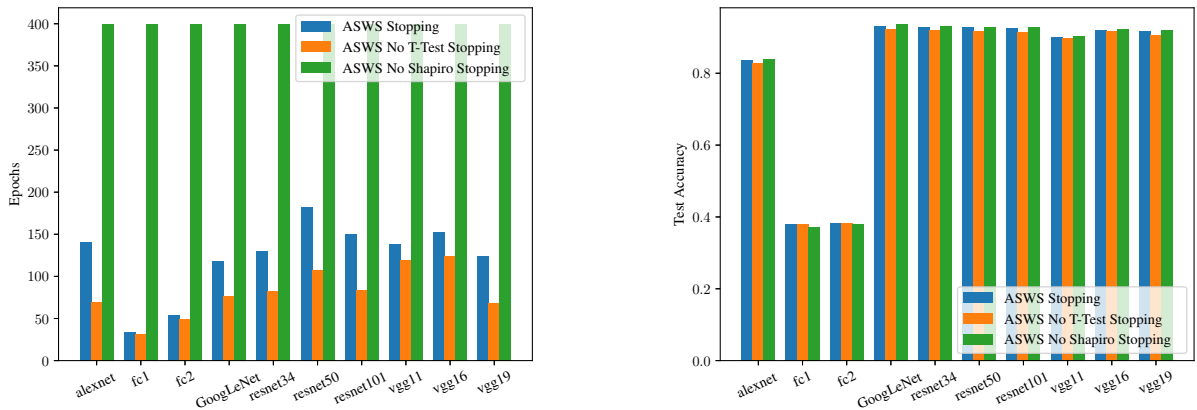

    \centering
    \begin{subfigure}{0.5\textwidth}
        \scalebox{0.5}{\input{figures/ASWTStandardCompAugment.pgf}}
    \end{subfigure}%
    \begin{subfigure}{0.5\textwidth}
        \centering
        \scalebox{0.5}{\input{figures/ASWTStandardCompAugmentAcc.pgf}}
    \end{subfigure}
    \caption{These plots compare the ASWS method when either no t-test is used or no Shapiro-Wilk Test is used, against the standard ASWS method. For each model, the average hyperparameters for ASWS were used and the Stopping Epoch and test accuracy were averaged across all 10 runs. ASWS has an average stopping epoch of 400, since each model was only trained for 400 epochs; ASWS with no Shapiro-Wilk test effectively never indicates to stop training. The Shapiro-Wilk test has a greater impact on the quality of ASWS, whereas the t-test seems to slightly improve the test accuracy.}
    \label{ASWTNoShapComp}
\end{figure*}

In normal ASWS, stopping is indicated when both the T-Test and Shapiro-Wilk tests have a significance greater than 0.97 for a certain number of the last epochs. \autoref{ASWTNoShapComp} compares the performance of ASWS without using the T-Test and Shapiro-Wilk test. When not using the Shapiro-Wilk test, the stopping epoch is 400; this is due to the fact that each model was only trained for 400 epochs, so without the Shapiro-Wilk test the model was never indicated to stop training. The T-Test indicates how close the mean of accuracy changes is to 0, which is unlikely to happen frequently enough to end training. Similarly, without the T-Test, ASWS indicates to stop earlier but at the cost of final test accuracy. Since the Shapiro-Wilk test tests normality, the distribution of test accuracy changes may be normally distributed but still overall positive, which means further training is possible.

\section{Automated Learning Rate Scheduling}
\label{sec:lr_scheduling}

Once optimization stops improving the neural network, from a naive perspective it's done learning. However, better performance can be achieved by decreasing the learning rate and fine tuning the model further, often multiple time. This is required to replicate state-of-the-art performance results for neural network classifiers. We show the intuitive result that using the ASWS as an automatic learning rate schedule (when the test says learning has stopped, decrease the learning rate by a factor of 10) is an effective method to automate learning rate scheduling. To do this we also introduce a $forcedEpochs$ parameter, where after adjusting the learning rate we train for $forcedEpochs$ before we perform the ASWS again. This gives learning time to restart after a drop in learning rate.

PyTorch natively includes several LR scheduling functions, of which we use three that we found performed the best. The StepLR scheduler multiplies the learning rate by some $\gamma \in [0,1]$ after $\mathsf{stepSize}$ epochs. Additionally, the ReduceLROnPLateau scheduler multiplies the learning rate by $\gamma$ if the test accuracy has not increased in $\mathsf{patience}$ epochs. Finally, we use the CyclicLR scheduler, which was introduced by \cite{smith2017cyclical}. The CyclicLR scheduler varies the learning between an upper and lower bound while training, which allows greater performance in fewer training epochs. The hyperparameters used for CyclicLR are the optimal hyperparameters for training GoogLeNet as determined by \cite{smith2017cyclical}. The hyperparameters used for the remaining scheduling functions were slightly hand-tuned from those provided by the PyTorch examples, and are included in \autoref{SchedulerParams}.

\begin{figure*}
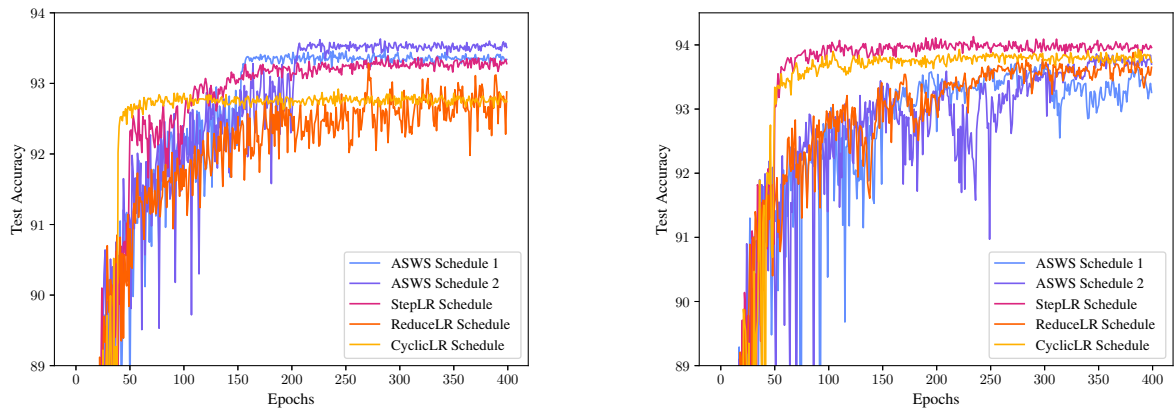

\begin{subfigure}{0.5\textwidth}
    \centering
    \scalebox{0.5}{\input{figures/resnet101scheduled6.pgf}}
\end{subfigure}%
\begin{subfigure}{0.5\textwidth}
    \centering
    \scalebox{0.5}{\input{figures/GoogLeNetscheduled6.pgf}}
    \end{subfigure}
    \caption{The figure on the left is ResNet101 models trained on CIFAR10, with different learning rate schedules. ASWS Models 1 and 2 used ASWS learning rate scheduling with 5 forced epochs after each learning rate change. The ASWS models eventually outperform the other schedules. On the right is GoogLeNet models trained on CIFAR10 with the same learning rate schedules. In this case, ASWS model2 roughly ties with the conventional schedules in terms of performance and convergence rate, but is surpassed throughout by the Cyclic and Step learning rate schedules}
    \label{fig:lr_schedule_1}
\end{figure*}

We experimentally compared two sets of ASWS hyperparameters, listed in \autoref{SchedulerParams}, against each of the three built in PyTorch functions described above on the ResNet101 and GoogLeNet models on the CIFAR10 dataset. For each scheduler, we perform 10 training runs per scheduler. The average max accuracy achieved is included in \autoref{LRSchedulerModelComp}. For the test accuracy curves shown in \autoref{fig:lr_schedule_1}, the model with highest test accuracy is plotted. As shown in \autoref{fig:lr_schedule_1}, for ResNet101 ASWS learning rate scheduling initially match other schedulers, but eventually outperforms them. When averaged over all 10 runs, the CyclicLR schedule slightly outperforms both of the ASWS schedules, as shown in \autoref{LRSchedulerModelComp}. For GoogLeNet, \autoref{fig:lr_schedule_1} shows that ASWS schedule 2 is roughly tied with the conventional schedulers in terms of end accuracy but is ultimately outperformed by the StepLR schedule and CyclicLR schedule. The ASWS scheduler hyperparameters, included in \autoref{SchedulerParams}, were chosen based on the optimized hyperparameter search results previously. However, the sample size $n$ was slightly increased so as to force the scheduler to wait more epochs before adjusting the schedule. Additionally, varying values for $\mathsf{slackProp}$ were used to examine the different effects on performance, since $\mathsf{slackProp}$ has the greatest variability as shown in \autoref{fig:hyperparameterseries}. Since $\mathsf{slackProp}$ has a significant effect on ASWS, it's expected that it will also have a significant effect on the LR scheduler. Note that the methods are all based on SGD except for vanilla ADAM

\section{Conclusion}


We introduce a novel statistical significance stopping criterion (Augmented Shapiro Wilk Stopping) to determine if a neural network is still learning the distribution during training. This allows practitioners to know when to automatically stop neural networks in a principled manner, in contrast to the heuristics typically used.
We show that ASWS appears to represent something of a happy medium in the space of stopping conditions. It stops in 77\% or less steps than all popular conditions but two without a meaningful loss in final accuracy. The two conditions that stops sooner than the ASWS does so as the expense of 2-4\% final accuracy even with tuned hyperparameters.
ASWS relies on three hyperparameters, and we show that tuning two is not of practical concern, while leaving the need to tune $SlackProp$ as a future work. We further show the ASWS learning rate scheduler can achieve comparable performance to schedulers which are commonly used in fewer iterations. All of these advances are of potentially great use towards a more environmentally sustainable machine learning, faster prototyping, and less far computational expense during large AutoML training endeavours

While our algorithm is in a state that practitioners can use it, there are still several logical extensions of this work to pursue. These include exploring the possibility of alternative smoothing methods and getting the test to function effectively on loss curves (allowing it to be applied more generally than for classification problems), exploring the performance on vast arrays of learning rates and optimizers (which we cannot exhaustively attempt given our computational resources), trying to extend the test to continuously modulate learning rate (turning it into a true line search), and using the test to try to achieve better model performance outright (instead of using it to try to stop the performance sooner).

We hope that this inspires more research into and usage of efficient and principled methods for stopping neural network training; this would be able to save very large amounts of time and resources in the wide scale training of neural networks in the future.

{\small
\bibliographystyle{icml2021}
\bibliography{main}
}

\onecolumn

\appendix


\begin{table}
\begin{center}
\begin{tabular}{ |c|c| } 
 \hline
    \textbf{ASWS Hyperparameter} & \textbf{Search Space} \\
 \hline
    $\gamma$ & [0.1, 0.8875] with step size of 0.1125 \\
 \hline
    n & [5, 19] with step size of 2 \\
 \hline
    slackProp & [0.05, 0.95] with step size of 0.082 \\
 \hline
\end{tabular}
\end{center}
\caption{ASWS Hyperparameter values used in grid search}
\label{GridSearchValues}
\end{table}

\begin{table}
\begin{center}
\begin{tabular}{ |c|c| } 
 \hline
    \textbf{Stopping Method Hyperparameter} & \textbf{Search Space} \\
 \hline
    $\mathsf{patience}$ & [1, 29] with step size of 2 \\
 \hline
    $\mathsf{window}$ & [25, 150] with step size of 25 \\
 \hline
    $\mathsf{min}$ $\delta $ & [0.001, 0.025] with step size of 0.004 \\
 \hline
    $\mathsf{min}$ $\delta $ $\mathsf{average}$ & [0.001, 0.025] with step size of 0.004 \\
 \hline
\end{tabular}
\end{center}
\caption{Heuristic Hyperparameter values used in grid search}
\label{HeuristicGridSearchValues}
\end{table}


\begin{table}
\begin{center}
\begin{tabular}{ |c|c|c|c|c|c| } 
 \hline
    \textbf{Model} & \textbf{Optimizer} & \textbf{Loss Function} & \textbf{LR} & \textbf{Momentum} & \textbf{LR Decay Rate} \\
 \hline
    AlexNet & SGD & Cross Entropy & 0.1 & 0 & 0\\
 \hline
    fc1 & SGD & Cross Entropy & 0.1 & 0 & 0\\
 \hline
    fc2 & SGD & Cross Entropy & 0.1 & 0 & 0\\
 \hline
    GoogLeNet & SGD & Cross Entropy & 0.1 & 0 & 0\\
 \hline
    ResNet34 & SGD & Cross Entropy & 0.1 & 0 & 0\\
 \hline
    ResNet50 & SGD & Cross Entropy & 0.1 & 0 & 0\\
 \hline 
    ResNet101 & SGD & Cross Entropy & 0.1 & 0 & 0\\ 
 \hline
    VGG11 & SGD & Cross Entropy & 0.1 & 0 & 0\\
 \hline
    VGG16 & SGD & Cross Entropy & 0.1 & 0 & 0\\
 \hline
    VGG19 & SGD & Cross Entropy & 0.1 & 0 & 0\\
 \hline
\end{tabular}
\end{center}
\caption{Hyperparameters used for training each neural network on CIFAR10. These hyperparameters were used for all 10 trials for each model.}
\end{table}

\begin{table}
\begin{center}
\begin{tabular}{ |c|c| } 
 \hline
    \textbf{Schedule} & \textbf{Hyperparameters} \\
 \hline
    StepLR & $\gamma$=0.5, $\mathsf{stepSize}$=50, $\mathsf{initlr}$=0.25 \\
 \hline
    CyclicLR & $\mathsf{mode}$=triangular2, $\mathsf{minLR}$=0.01, $\mathsf{maxLR}$=0.026, $\mathsf{stepsize}$=30000 \\
 \hline
    ReduceLR & $\gamma$=0.1, $\mathsf{patience}$=20, $\mathsf{initlr}$=0.25 \\
 \hline
    Resnet-ASWS Schedule 1 & $\gamma$=0.60, $n$=17, $\mathsf{slackProp}$=0.35, $\mathsf{initlr}$=0.1\\
 \hline
    Resnet-ASWS Schedule 2 & $\gamma$=0.60, $n$=17, $\mathsf{slackProp}$=0.05, $\mathsf{initlr}$=0.1\\
 \hline
    GoogLeNet-ASWS Schedule 1 & $\gamma$=0.76, $n$=19, $\mathsf{slackProp}$=0.05, $\mathsf{initlr}$=0.1\\
 \hline
    GoogLeNet-ASWS Schedule 2 & $\gamma$=0.76, $n$=19, $\mathsf{slackProp}$=0.35, $\mathsf{initlr}$=0.1\\
\hline
\end{tabular}
\end{center}
\caption{Learning Rate Schedule Hyperparameters}
\label{SchedulerParams}
\end{table}

\begin{table}
\begin{center}
\begin{tabular}{ |c|c| } 
 \hline
    \textbf{Stopping Method} & \textbf{Hyperparameters} \\
 \hline
    Performance Stopping & $\mathsf{K}$=60 \\
 \hline
    Patience Stopping & $\mathsf{patience}$=3 \\
 \hline
    Minimum Diff Stopping & $\mathsf{patience}$=27, $\mathsf{min}$  $\delta$=0.013 \\
 \hline
    Average Diff Stopping & $\mathsf{window}$=150, $\mathsf{min}$ $\delta $ $\mathsf{average}$=0.001\\
 \hline

\end{tabular}
\end{center}
\caption{Performance Stopping Method Hyperparameters}
\label{StandardStopParams}
\end{table}

\begin{table}
\begin{center}
\begin{tabular}{ |c|c|c| } 
 \hline
    \textbf{LR Scheduler} & \textbf{ResNet101 Average Accuracy} & \textbf{GoogLeNet Average Accuracy} \\
 \hline
    ASWT 1 & 93.0\% & 93.3\%\\
 \hline
    ASWT 2 Stopping & 93.1\%& 93.5\% \\
 \hline
    StepLR & 92.9\% & 93.9\%\\
 \hline
    ReduceLR & 92.5\% & 93.6\%\\
 \hline
    CyclicLR & 93.2\% & 93.7\%\\
 \hline
\end{tabular}
\end{center}
\caption{Comparison of LR Scheduler Results}
\label{LRSchedulerModelComp}
\end{table}

\begin{table}
\begin{center}
\begin{tabular}{ |c|c|c|c|c|c|c|c|c|c|c| } 
\hline
Model & AvgStdEpoch & AvgASWTEpoch & AvgPatEpoch & AvgMindEpoch & AvgAvgesEpoch \\
\hline
AlexNet & 216.3 & 127.5 & 92.5 & 78.1 & 159.8 \\
\hline
FC1 & 69.9 & 70.6 & 47.7 & 118.4 & 150 \\
\hline
FC2 & 133 & 142.9 & 38.7 & 102.1 & 150  \\
\hline
GoogLeNet & 202.4 & 88.3 & 60.7 & 111.3 & 154.5 \\
\hline
ResNet34 & 187.5 & 130.1 & 73.3 & 112.2 & 156.5 \\
\hline
ResNet50 & 251.2 & 142.1 & 47.2 & 104.8 & 156.6 \\
\hline
ResNet101 & 211.1 & 129.1 & 45.5 & 90.3 & 156.4 \\
\hline
VGG11 & 222.4 & 145.8 & 54.1 & 105.9 & 155.3 \\
\hline
VGG16 & 238.8 & 119.6 & 60.5 & 97.1 & 155.3 \\
\hline
VGG19 & 204.6 & 99.4 & 68.6 & 99.1 & 156.5 \\
\hline
\end{tabular}
\end{center}
\caption{ASWS Stopping Epoch compared to other stopping criterion}
\label{ASWTHeuristicCompEochs}
\end{table}

\begin{table}
\begin{center}
\begin{tabular}{ |c|c|c|c|c|c|c|c|c|c|c| } 
\hline
Model & AvgStdAcc & AvgASWTAcc & AvgPatAcc & AvgMindAcc & AvgAvgesAcc \\
\hline
AlexNet  & 0.84208 & 0.83541 & 0.831 & 0.83026 & 0.83818 \\
\hline
FC1  & 0.38507 & 0.38212 & 0.38161 & 0.24664 & 0.38554 \\
\hline
FC2  & 0.38845 & 0.38588 & 0.38074 & 0.38587 & 0.38808 \\
\hline
GoogLeNet  & 0.93519 & 0.92719 & 0.90409 & 0.92942 & 0.93292 \\
\hline
ResNet34  & 0.93335 & 0.92799 & 0.9109 & 0.92716 & 0.93153 \\
\hline
ResNet50 & 0.93245 & 0.92497 & 0.90384 & 0.92314 & 0.92746 \\
\hline
ResNet101  & 0.93148 & 0.92549 & 0.90031 & 0.92268 & 0.9285 \\
\hline
VGG11  & 0.90563 & 0.89793 & 0.86208 & 0.89888 & 0.90236 \\
\hline
VGG16  & 0.9241 & 0.91695 & 0.90558 & 0.91587 & 0.92056 \\
\hline
VGG19 & 0.92234 & 0.91438 & 0.90534 & 0.91518 & 0.91989 \\
\hline
\end{tabular}
\end{center}
\caption{ASWS Test Accuracy compared to other stopping criterion}
\label{ASWTHeuristicCompAcc}
\end{table}

\end{document}